\documentclass{article}
\usepackage{spconf,amsmath,graphicx,hyperref}
\usepackage{enumitem}
\usepackage{multirow}
\usepackage{floatrow}
\usepackage{xcolor}
\usepackage{colortbl}
\usepackage{booktabs}

\usepackage{caption}
\usepackage{subcaption}
\usepackage{balance,cite}
\definecolor{cvprblue}{rgb}{0.21,0.49,0.74}
\definecolor{silver}{gray}{0.95}
\definecolor{lightgray}{gray}{0.98}
\definecolor{lightred}{rgb}{0.99,0.94,0.95}
\definecolor{lightblue}{rgb}{0.94,0.95,0.99}
\definecolor{lightgreen}{rgb}{0.94,0.99,0.95}
\definecolor{best}{rgb}{0.62,0.58,0.99}
\definecolor{secondbest}{rgb}{0.84,0.83,0.99}

\newfloatcommand{capbtabbox}{table}[][\FBwidth]

\newcommand\blfootnote[1]{%
  \begingroup
  \renewcommand\thefootnote{}\footnote{#1}%
  \addtocounter{footnote}{-1}%
  \endgroup
}
\title{Linking Faces and Voices Across Languages:\\ Insights from the FAME 2026 Challenge}

\name{
\begin{tabular}{c}
 Marta Moscati$^{1}$\textsuperscript{\textdagger}, Ahmed Abdullah$^{2}$\textsuperscript{\textdagger}, Muhammad Saad Saeed$^{3}$\textsuperscript{\textdagger}, Shah Nawaz$^{1}$\textsuperscript{\textdagger},\\ Rohan Kumar Das$^{4}$ 
 \textsuperscript{\textdagger},  Muhammad Zaigham Zaheer$^{5}$,  Junaid Mir$^{6}$,\\ Muhammad Haroon Yousaf$^{6}$, Khalid Mahmood Malik$^{3}$, Markus Schedl$^{1,7}$  \\
\end{tabular}}
\address{
$^{1}$Johannes Kepler University Linz, Austria, 
$^{2}$National University of Computer and Emerging Sciences, Pakistan\\
$^{3}$University of Michigan, USA, 
$^{4}$Fortemedia Singapore, Singapore \\
$^{5}$Mohamed bin Zayed University of Artificial Intelligence, United Arab Emirates \\
$^{6}$University of Engineering and Technology Taxila, Pakistan,\\
$^{7}$Human-centered AI Group, AI Lab, Linz Institute of Technology, Austria \\
\tt mavceleb@gmail.com
}

\begin{document}
%\ninept
%
\maketitle
\begin{abstract}
    % \blfootnote{\textsuperscript{\textdagger}Equal Contribution.}
    Over half of the world's population is bilingual and people often communicate under multilingual scenarios. The Face-Voice Association in Multilingual Environments (FAME) 2026 Challenge, held at ICASSP 2026, focuses on developing methods for face-voice association % and on analyzing the impact of multiple languages on the verification process
    that are effective when the language at test-time is different than the training one. This report provides a brief summary of the challenge.
\end{abstract}
%\
\vspace{-2mm}

\section{Introduction}
    \label{sec:intro}

\begin{figure}
    \centering
    \includegraphics[width=0.80\linewidth]{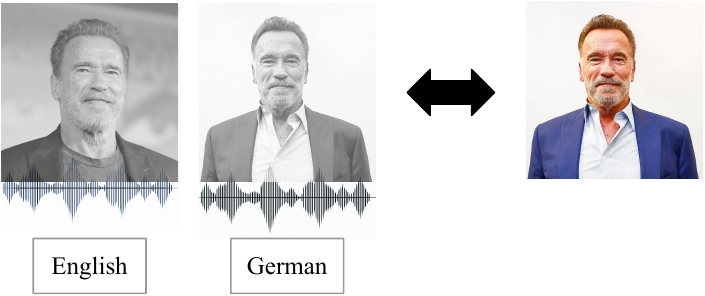}
    \vspace{2mm}
    \caption{
    The FAME 2026 Challenge overview: cross-modal verification to analyze the impact of multiple languages.}
    \label{fig:verification+protocol}
    \vspace{-4mm}
\end{figure}

    Humans can associate the voices and faces of people because the neuro-cognitive pathways for both modalities share the same structure~\cite{kamachi2003putting}. Inspired by this insight,
    Nagrani et al. have leveraged deep learning methods to establish an association between voices and faces for cross-modal speaker verification and matching tasks~\cite{nagrani2018seeing,nagrani2018learnable}. Since then, the task of correctly associating the face and voice of a speaker has received notable research interest~\cite{horiguchi2018face,nawaz2019deep,tao20b_interspeech,chen2023local,saeed2023single,shah2023speaker,hannan2025paeff}. 
    %Though previous studies have established association between faces and voices, the effect of multiple languages has not been explored. 
    As over half of the world population is bilingual, it is important to investigate the impact of language on face-voice %(F-V) acronym never used
    association. 
    As displayed in Fig.~\ref{fig:verification+protocol}, the Face-Voice Association in Multilingual Environments (FAME) 2026 Challenge aims to promote the development of cross-modal verification algorithms that are effective for multilingual speakers.  
\begin{NoHyper}
\blfootnote{\textsuperscript{\textdagger}Equal Contribution.}
\end{NoHyper}

\section{The Grand Challenge: FAME 2026}
    \noindent \textbf{Overview.} The task of the grand challenge is cross-modal speaker verification, where the goal is to verify whether the audio segment and a face image belong to the same speaker. In our challenge each speaker is represented while speaking more than one language. In particular,  using the samples of MAV-Celeb~\cite{nawaz2021cross,saeed2024synopsis}, we curated \textit{unheard} test sets, in which the language spoken by the speakers was never represented in the training data. This allowed us to evaluate whether the proposed solutions were subject to a performance deterioration under a change of language.

    \noindent \textbf{Baseline Method \& Starter Kit.}
     The baseline approach is named FOP and consists of a two-branch network that takes as input the embeddings of face and voice. The embeddings for the first branch are obtained using a convolutional neural network pre-trained on a large-scale facial recognition dataset~\cite{parkhi2015deep}.  
     The embeddings of the other branch are extracted using an audio encoding network~\cite{xie2019utterance} trained using the \textit{heard} language. 
    The network combines the complementary information from both modalities in a fused embedding and imposes orthogonal constraints on the two modalities to learn joint representations that represent distinct speakers in different ways. More information is available in the prior work on the baseline~\cite{saeed2022fusion} and the repository: \href{https://github.com/mavceleb/mavceleb_baseline}{https://github.com/mavceleb/mavceleb\_baseline}.

    \noindent \textbf{Dataset.} Following the first edition of the FAME Grand Challenge hosted at ACM Multimedia 2024~\cite{saeed2024face,saeed2024synopsis}, we updated the dataset including German as  additional language, as well as bilingual speakers. We curated a new dataset split consisting of $58$ English-German bilingual speakers. 
    The split provides language annotations, which allow to analyze the impact of multiple languages on face-voice association. Analogously to the previous splits, the samples are obtained from YouTube videos and consist of celebrities appearing in interviews, talk shows, and television debates~\cite{nawaz2021cross}. 
    The visual data spans a vast range of setups, including different poses, motion blurs, background clutters, video qualities, occlusions, and lighting conditions. Moreover, since the videos originate from real-world situations, they reproduce the same challenges that are encountered when deploying face-voice association tools in real-world scenarios, such as noise, background chatter or music, overlapping voices, and compression artifacts. These aspects render the dataset both challenging for existing algorithms, and useful for developing algorithms that can have an impact on real applications. 
    
    \noindent \textbf{Evaluation Plan.} Comprehensive details on baseline, metric, submission portal, and rules are provided in~\cite{moscati2025face}.

 \begin{table}[b!]
 \vspace{-2mm}
\caption{Performance comparison of top 5 teams with baseline in EER (\%).}
\centering
\setlength{\tabcolsep}{25pt}
\scalebox{0.80}{
\begin{tabular}{c l c}
\toprule
\textbf{Rank}  & \textbf{Team} & \textbf{EER (\%)} \\
\midrule
\rowcolor{lightred}
-  & Baseline~\cite{saeed2022fusion}     & 41.57 \\
\midrule
1 & Simicch~\cite{simic2025shared}      & 23.99 \\
2 & Areffarhadi~\cite{farhadipour2025towards}  & 24.73 \\
3 & Alpha\_code~\cite{hannan2025rfop}  & 33.11 \\
4 & LTINI~\cite{Zhang_2025}        & 33.18 \\
5 & Punkmale~\cite{fang2025xm}     & 33.51 \\
\bottomrule
\end{tabular}
}
\label{tab:eer}
\end{table}

    \begin{figure}[!t]
     \centering
     \begin{subfigure}[b]{0.98\textwidth}
         \centering
         \includegraphics[width=\textwidth]{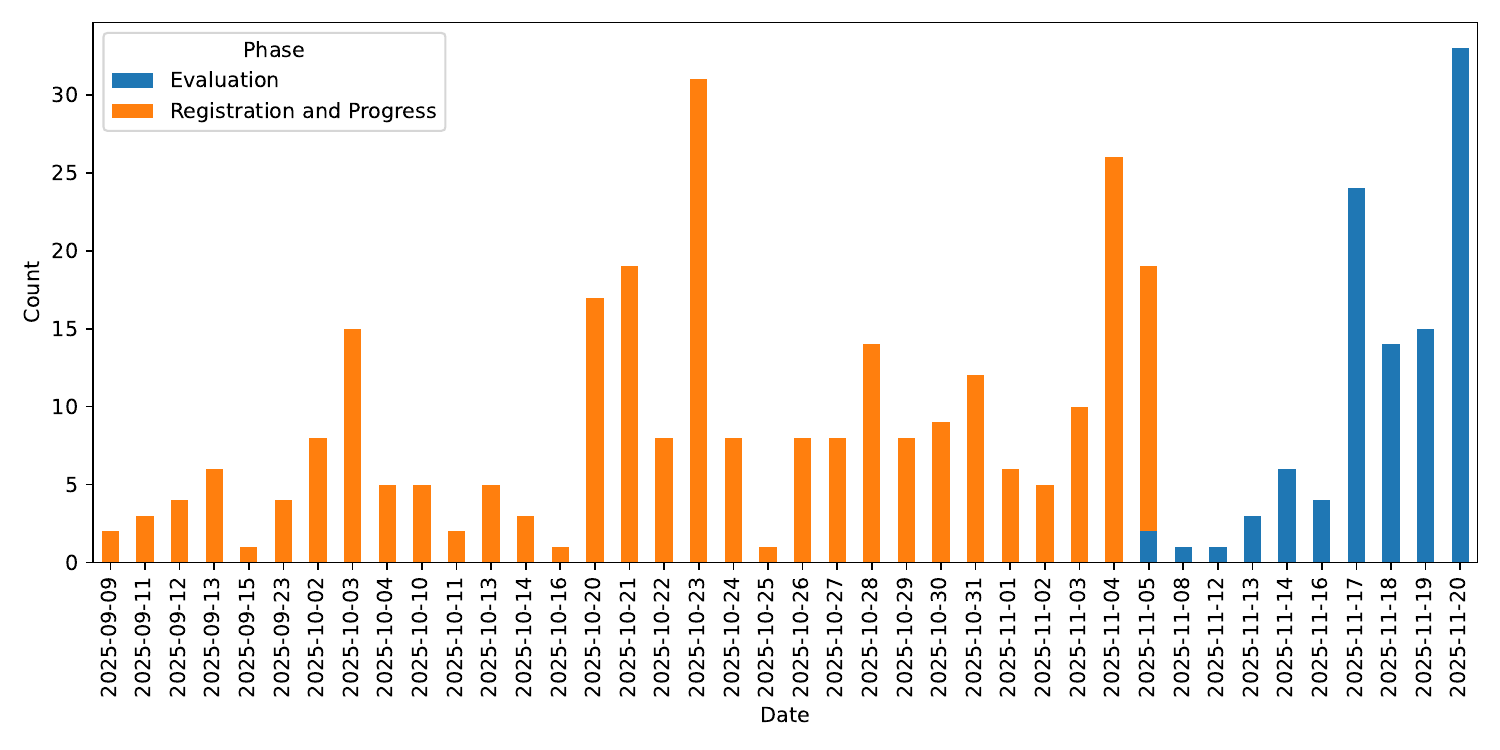}
         \caption{Daily submissions.}
         \label{fig:daily_submissions}
     \end{subfigure}
     % \hfill
     \begin{subfigure}[b]{0.98\textwidth}
         \centering
         \includegraphics[width=\textwidth]{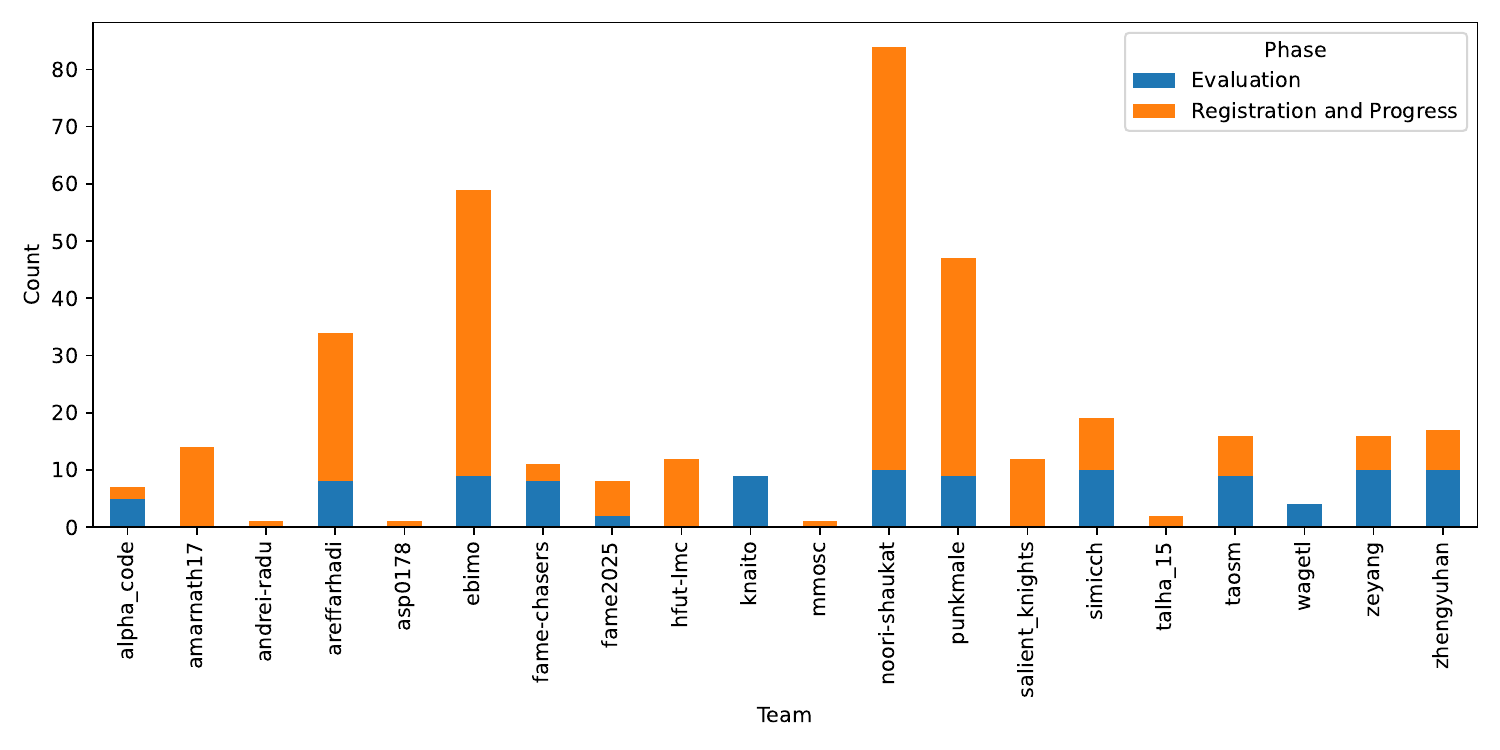}
         \caption{Team submissions.}
         \label{fig:stacked_graph}
     \end{subfigure}

          \begin{subfigure}[b]{0.98\textwidth}
         \centering
         \includegraphics[width=\textwidth]{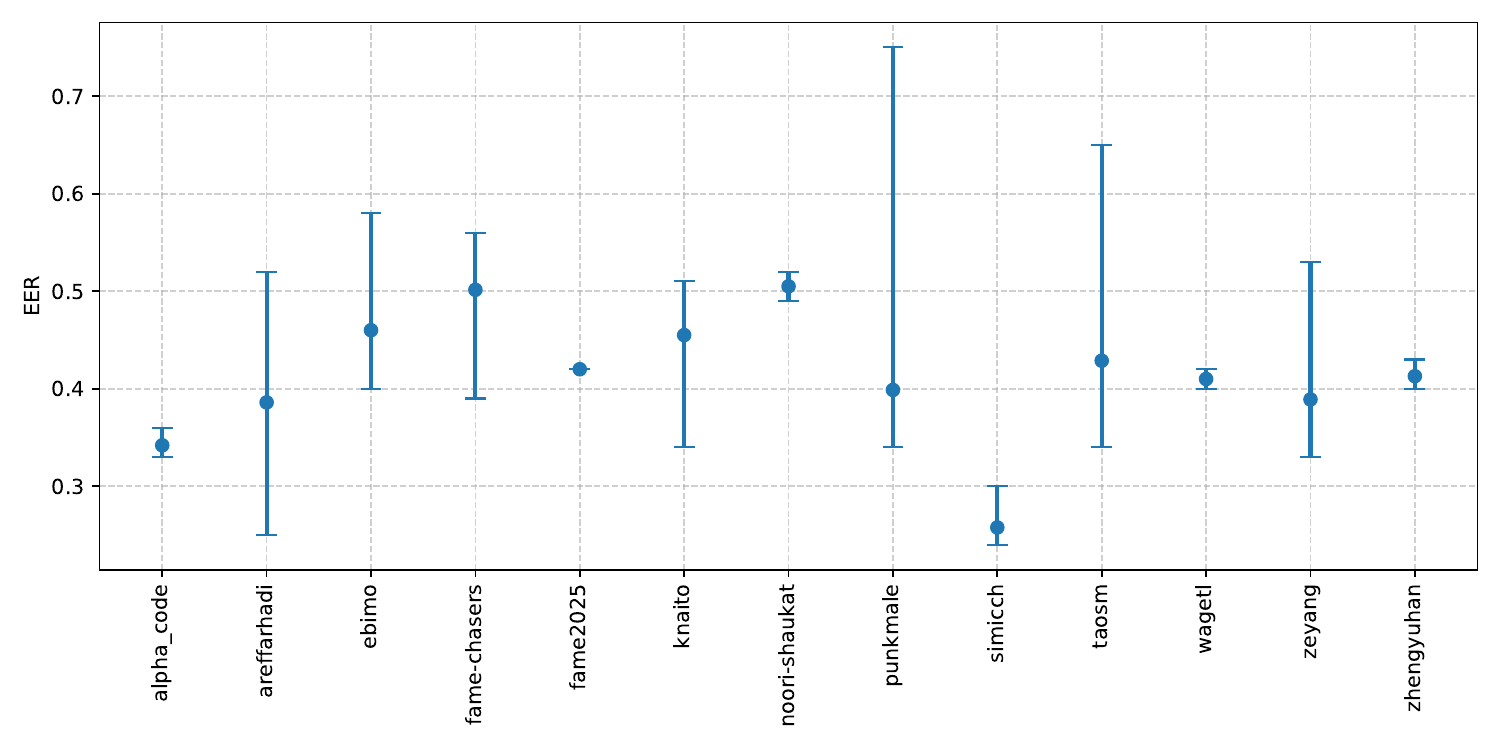}
         \vspace{-2mm}
         \caption{Team-wise mean EER.}
         \label{fig:eer_graph}
     \end{subfigure}
      \caption{(a) Overview of participant activity across grand challenge phases, showing the temporal distribution of submissions. (b) Aggregation of team submissions, emphasizing differences in total participation and the phase allocation between Registration/Progress and Evaluation phase. (c) Evaluation phase performance is summarized by team-wise mean EER, with error bars indicating standard deviation.}
        \label{fig:three graphs}
        \vspace{-2mm}
\end{figure}

    \noindent \textbf{Challenge Timeline and Results.} The challenge consisted of two phases. In the first phase (August $15$, $2025$, to November $20$, $2025$) participants' approaches were evaluated on the development set; each team was allowed to submit $10$ submissions per day, with a maximum number of $100$ submissions during the entire phase. In the second phase (November $05$, $2025$ to November $20$, $2025$) participants' approaches were evaluated on the unseen evaluation set; each team was allowed for a maximum of $10$ submissions over the whole phase.
    Fig.~\ref{fig:daily_submissions} shows the daily submission activity across all teams during the Progress and Evaluation phases, indicating distinct spikes that correspond to periods of increased engagement. Fig.~\ref{fig:stacked_graph} further breaks down these contributions by team, highlighting variations in participation intensity across the two phases. Finally, Fig.~\ref{fig:eer_graph} shows the team-wise performance during the Evaluation phase, expressed as mean Equal Error Rate (EER) with standard deviation error bars.
    Finally, Table~\ref{tab:eer} presents the EER for the top-5 teams in comparison to the baseline method~\cite{saeed2022fusion}. The participating teams demonstrated a notable improvement over the baseline method, with the leading team, Simicch~\cite{simic2025shared}, achieving an EER of $23.99$, a considerable reduction from the baseline EER of $41.57$. 
    This substantial performance gap underscores the success of the challenge, which led to the development of effective methodologies that advanced the state-of-the-art for face-voice association in multilingual environments.

\section{Acknowledgments}
This research was funded in whole or in part by the Austrian Science Fund (FWF): \url{https://doi.org/10.55776/COE12} as part of the Cluster of Excellence \href{https://www.bilateral-ai.net/}{Bilateral Artificial Intelligence}.

\balance
\bibliographystyle{IEEEbib}
%\small
\bibliography{strings,refs}

@article{Zhang_2025,
title={Contrastive Gated Fusion for Multilingual Speaker Verification},
url={http://dx.doi.org/10.36227/techrxiv.176620835.51369097/v1},
DOI={10.36227/techrxiv.176620835.51369097/v1},
publisher={Institute of Electrical and Electronics Engineers (IEEE)},
author={Zhang, Zeyang and Naito, Katsuhiko and Dahmani, Hajer},
year={2025},
month=dec }

@article{hannan2025rfop,
  title={{RFOP}: Rethinking Fusion and Orthogonal Projection for Face-Voice Association},
  author={Hannan, Abdul and Malik, Furqan and Jabbar, Hina and Sadiq, Syed Suleman and Noman, Mubashir},
  journal={arXiv preprint arXiv:2512.02860},
  year={2025}
}

@article{moscati2025face,
  title={Face-voice Association in Multilingual Environments {(FAME)} 2026 Challenge Evaluation Plan},
  author={Moscati, Marta and Abdullah, Ahmed and Saeed, Muhammad Saad and Nawaz, Shah and Das, Rohan Kumar and Zaheer, Muhammad Zaigham and Mir, Junaid and Yousaf, Muhammad Haroon and Malik, Khalid and Schedl, Markus},
  journal={arXiv preprint arXiv:2508.04592},
  year={2025}
}

@article{saeed2024face,
  title={Face-voice Association in Multilingual Environments {(FAME)} Challenge 2024 Evaluation Plan},
  author={Saeed, Muhammad Saad and Nawaz, Shah and Tahir, Muhammad Salman and Das, Rohan Kumar and Zaheer, Muhammad Zaigham and Moscati, Marta and Schedl, Markus and Khan, Muhammad Haris and Nandakumar, Karthik and Yousaf, Muhammad Haroon},
  journal={arXiv preprint arXiv:2404.09342},
  year={2024}
}

@inproceedings{tao20b_interspeech,
  author={Ruijie Tao and Rohan Kumar Das and Haizhou Li},
  title={{Audio-Visual Speaker Recognition with a Cross-Modal Discriminative Network}},
  year=2020,
  booktitle={Interspeech},
}

@article{fang2025xm,
  title={{XM-ALIGN}: Unified Cross-Modal Embedding Alignment for Face-Voice Association},
  author={Fang, Zhihua and Tao, Shumei and Wang, Junxu and He, Liang},
  journal={arXiv preprint arXiv:2512.06757},
  year={2025}
}

@article{farhadipour2025towards,
  title={Towards Language-Independent Face-Voice Association with Multimodal Foundation Models},
  author={Farhadipour, Aref and Vukovic, Teodora and Dellwo, Volker},
  journal={arXiv preprint arXiv:2512.02759},
  year={2025}
}

@article{simic2025shared,
  title={Shared Multi-modal Embedding Space for Face-Voice Association},
  author={Simic, Christopher and Riedhammer, Korbinian and Bocklet, Tobias},
  journal={arXiv preprint arXiv:2512.04814},
  year={2025}
}

@inproceedings{xie2019utterance,
  title={Utterance-level aggregation for speaker recognition in the wild},
  author={Xie, Weidi and Nagrani, Arsha and Chung, Joon Son and Zisserman, Andrew},
  booktitle={ICASSP},
  year={2019},
}

@inproceedings{nawaz2019deep,
  title={Deep latent space learning for cross-modal mapping of audio and visual signals},
  author={Nawaz, Shah and Janjua, Muhammad Kamran and Gallo, Ignazio and Mahmood, Arif and Calefati, Alessandro},
  booktitle={DICTA},
  year={2019},
}

@inproceedings{parkhi2015deep,
  title={Deep face recognition},
  author={Parkhi, Omkar and Vedaldi, Andrea and Zisserman, Andrew},
  booktitle={BMVC},
  year={2015},
}

@inproceedings{shah2023speaker,
  title={Speaker recognition in realistic scenario using multimodal data},
  author={Shah, Saqlain Hussain and Saeed, Muhammad Saad and Nawaz, Shah and Yousaf, Muhammad Haroon},
  booktitle={ICAI},
  year={2023},
}

@inproceedings{chen2023local,
  title={Local-Global Contrast for Learning Voice-Face Representations},
  author={Chen, Guangyu and Zhang, Deyuan and Liu, Tao and Du, Xiaoyong},
  booktitle={ICIP},
  year={2023},
}

@inproceedings{horiguchi2018face,
  title={Face-voice matching using cross-modal embeddings},
  author={Horiguchi, Shota and Kanda, Naoyuki and Nagamatsu, Kenji},
  booktitle={ACM Multimedia},
  year={2018}
}

@inproceedings{nawaz2021cross,
  title={Cross-modal speaker verification and recognition: A multilingual perspective},
  author={Nawaz, Shah and Saeed, Muhammad Saad and Morerio, Pietro and Mahmood, Arif and Gallo, Ignazio and Yousaf, Muhammad Haroon and Del Bue, Alessio},
  booktitle={CVPR Workshops},
  year={2021}
}

@inproceedings{saeed2024synopsis,
  title={A Synopsis of {FAME 2024} Challenge: Associating Faces with Voices in Multilingual Environments},
  author={Saeed, Muhammad Saad and Nawaz, Shah and Moscati, Marta and Das, Rohan Kumar and Tahir, Muhammad Salman and Zaheer, Muhammad Zaigham and Liaqat, Muhammad Irzam and Khan, Muhammad Haris and Nandakumar, Karthik and Yousaf, Muhammad Haroon and others},
  booktitle={ACM Multimedia},
  year={2024},
}

@inproceedings{saeed2023single,
  title={Single-branch network for multimodal training},
  author={Saeed, Muhammad Saad and Nawaz, Shah and Khan, Muhammad Haris and Zaheer, Muhammad Zaigham and Nandakumar, Karthik and Yousaf, Muhammad Haroon and Mahmood, Arif},
  booktitle={ICASSP},
  year={2023},
}

@inproceedings{nagrani2018seeing,
  title={Seeing voices and hearing faces: Cross-modal biometric matching},
  author={Nagrani, Arsha and Albanie, Samuel and Zisserman, Andrew},
  booktitle={CVPR},
  year={2018}
}

@inproceedings{nagrani2018learnable,
  title={Learnable pins: Cross-modal embeddings for person identity},
  author={Nagrani, Arsha and Albanie, Samuel and Zisserman, Andrew},
  booktitle={ECCV},
  year={2018}
}

@article{kamachi2003putting,
  title={Putting the face to the voice': Matching identity across modality},
  author={Kamachi, Miyuki and Hill, Harold and Lander, Karen and Vatikiotis-Bateson, Eric},
  journal={Current Biology},
  volume={13},
  number={19},
  year={2003},
  publisher={Elsevier}
}

@inproceedings{hannan2025paeff,
  title={{PAEFF}: Precise Alignment and Enhanced Gated Feature Fusion for Face-Voice Association},
  author={Hannan, Abdul and Manzoor, Muhammad Arslan and Nawaz, Shah and Liaqat, Muhammad Irzam and Schedl, Markus and Noman, Mubashir},
booktitle={Interspeech},
year={2025},
}

@inproceedings{saeed2022fusion,
  title={Fusion and orthogonal projection for improved face-voice association},
  author={Saeed, Muhammad Saad and Khan, Muhammad Haris and Nawaz, Shah and Yousaf, Muhammad Haroon and Del Bue, Alessio},
  booktitle={ICASSP},
  year={2022},
}

\end{document}